\documentclass[10pt,twocolumn,letterpaper]{article}

\usepackage{wacv}
\usepackage{times}
\usepackage{epsfig}
\usepackage{graphicx}
\usepackage{amsmath}
\usepackage{amssymb}

\usepackage{graphicx} 
\usepackage{subfigure}
\usepackage{flushend}

\usepackage{amsmath} 
\usepackage{amssymb}  
\usepackage{ulem}
\usepackage{hyperref}

\usepackage[table]{xcolor}
\usepackage{breqn}
\usepackage{booktabs}
\usepackage{fancyhdr}

\definecolor{wrongultramarine}{rgb}{0.07,0.04,0.56}
\definecolor{roadclass}{rgb}{110, 175,30}

\definecolor{bicycle}{rgb}{0.47,0.04,0.13}

\newcommand{\cmmnt}[1]{\ignorespaces}

\wacvfinalcopy 

\def\wacvPaperID{0075} 

\ifwacvfinal\fi
\begin{document}

\title{Unbiasing Semantic Segmentation For Robot Perception using Synthetic Data Feature Transfer}

\author{Jonathan C Balloch \hspace{2cm} Varun Agrawal \hspace{2cm} Irfan Essa \hspace{2cm} Sonia Chernova \\
Georgia Institute of Technology \\
{\tt\small balloch@gatech.edu}
}

\maketitle
\ifwacvfinal\thispagestyle{empty}\fi

\begin{abstract}
Robot perception systems need to perform reliable image segmentation in real-time on noisy, raw perception data. State-of-the-art segmentation approaches use large CNN models and carefully constructed datasets; however, these models focus on accuracy at the cost of real-time inference. Furthermore, the standard semantic segmentation datasets are not large enough for training CNNs without augmentation and are not representative of noisy, uncurated robot perception data. We propose improving the performance of real-time segmentation frameworks on robot perception data by transferring features learned from synthetic segmentation data. We show that pretraining real-time segmentation architectures with synthetic segmentation data instead of ImageNet improves fine-tuning performance by reducing the bias learned in pretraining and closing the \textit{transfer gap} as a result. Our experiments show that our real-time robot perception models pretrained on synthetic data outperform those pretrained on ImageNet for every scale of fine-tuning data examined. Moreover, the degree to which synthetic pretraining outperforms ImageNet pretraining increases as the availability of robot data decreases, making our approach attractive for robotics domains where dataset collection is hard and/or expensive.


\end{abstract}

\section{Introduction} \label{sec:num1}

Intelligent robots depend on reliable semantic segmentation of objects to support effective physical interactions. While convolutional neural networks (CNNs) have significantly improved the performance of segmentation solutions, two challenges remain. First, the pixel-level classification task of segmentation requires models that can represent more complex distributions than those of object detection with bounding box regression and image classification. Second, the cost of annotating pixel-level segmentation data is prohibitive at the scale needed to train CNNs. Furthermore, many state-of-the-art semantic segmentation approaches do not readily extend to robot vision, as models with millions of parameters requiring billions of operations to classify each image remain impractical \cite{canziani2016analysis}. Robot vision requires efficient CNN segmentation architectures that can be successfully trained on uncurated datasets acquired from robots interacting with the world, and segment inherently noisy perception data in real-time.

To this end, we present a novel perspective for developing robust semantic segmentation systems for robot perception. Specifically, our approach of transferring representations pretrained on synthetic segmentation data to real-time perception systems strictly improves performance by reducing bias in training. Synthetic segmentation datasets have the advantages of being scalable to millions of examples, giving perfect ground truth labels without extra annotation effort, and having less dataset bias. 



Recent work has demonstrated that for large, non-real-time segmentation architectures, models pretrained with synthetic datasets can out-perform models pretrained on ImageNet \cite{mccormac2017beatimagenet}. For our task aimed at robot vision, we explore whether this result holds for real-time segmentation architectures with less representational capacity by pretraining a real-time model with synthetic data \cite{mccormac2016scenenet} and comparing its fine-tuned performance with a model pretrained on ImageNet data. Our results to this ablation experiment show that for real-time segmentation architectures, synthetic data pretraining yields better performance than ImageNet, and that these performance improvements are greater than the performance gains in larger architectures \cite{mccormac2017beatimagenet}.



To further validate our hypothesis, we investigate how models that are pretrained with synthetic data handle noise and scale in robot datasets. Typically datasets acquired from robots have sparse supervision, making it difficult to train a semantic segmentation model that accommodates the increased noise and bias in both the inputs and the labels. We examine the performance of synthetic data pretrained models (in comparison to ImageNet pretrained models) by fine-tuning on various subsets of a robot navigation dataset \cite{ruiz2017robot} to quantify the benefits as the amount of supervised fine-tuning data is decreased. Our results show the synthetic data pretrained models outperform the models with ImageNet pretraining for every amount of fine-tuning data, and that the performance improvement increases as the number of robot data training samples decreases. 

Lastly, we consider whether the improvements in performance due to synthetic data pretraining is caused strictly by similarity of the high-level scenarios of the synthetic pretraining data and the target data. To test this, we consider the two standard datasets from the ablation experiment as our target data, where the high-level scenarios are ``driving'' and ``indoor navigation'' respectively. For both of these target datasets sets we train two models, one pretrained on a dataset designed for a similar high-level scenario and the other pretrained on data from a different high-level scenario. Our results show that models pretrained using a synthetic dataset, that is similar to the target task, display improvements on the task. However, we also show that there is a greater benefit for the target robot vision task in pretraining on a synthetic dataset that has more input diversity, more coverage, and less bias, regardless of the high-level similarity.

Our primary contributions are: (1) Extending the benefits of transferring features from synthetic data pretraining instead of ImageNet pretraining to real-time-optimized segmentation CNNs, (2) Demonstrating that transferring features from synthetic segmentation data helps reduce the amount of target robot data needed for strong real-time segmentation performance, and (3) Exploring the effect of ``high-level domain similarity'' between datasets, and showing that while synthetic data that has a similar high-level domain does give some improvement to performance, other properties of data, such as scale, diversity, and bias, have a greater effect on performance.

\begin{figure}[ht]
\centering
    (a) \subfigure{\includegraphics[width = 0.3\columnwidth]{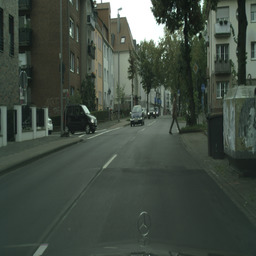}} 
    \subfigure{\includegraphics[width = 0.3\columnwidth]{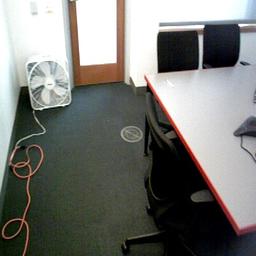}} 
    \subfigure{\includegraphics[width = 0.3\columnwidth]{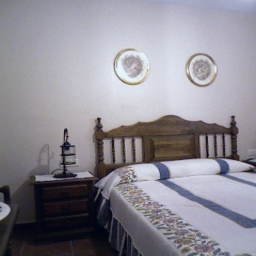}}  \\ 
    (b) \subfigure{\includegraphics[width = 0.3\columnwidth]{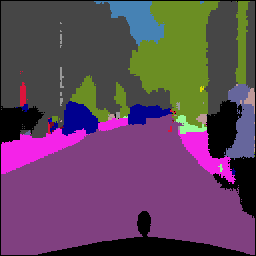}} 
    \subfigure{\includegraphics[width = 0.3\columnwidth]{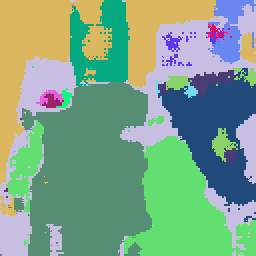}} 
    \subfigure{\includegraphics[width = 0.3\columnwidth]{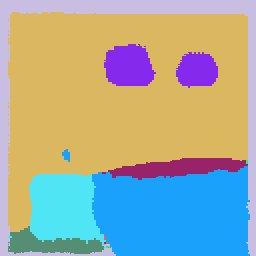}} \\ 
    (c) \subfigure{\includegraphics[width = 0.3\columnwidth]{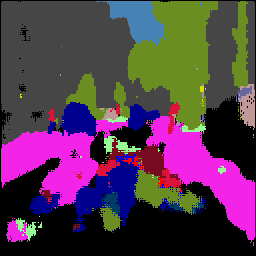}} 
    \subfigure{\includegraphics[width = 0.3\columnwidth]{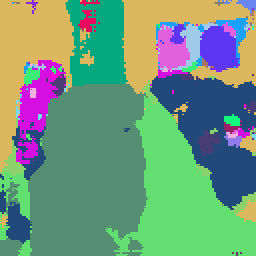}} 
    \subfigure{\includegraphics[width = 0.3\columnwidth]{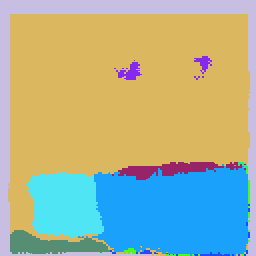}} \\
    (d) \subfigure{\includegraphics[width = 0.3\columnwidth]{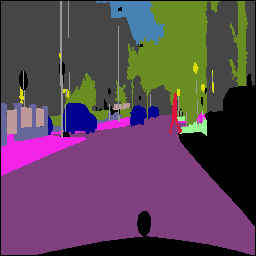}} 
    \subfigure{\includegraphics[width = 0.3\columnwidth]{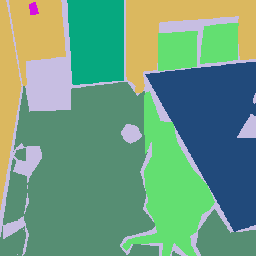}} 
    \subfigure{\includegraphics[width = 0.3\columnwidth]{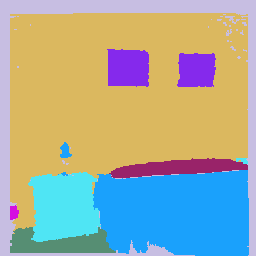}} \\ 

\caption{Qualitative results from the ablation and training set reduction experiments. Cityscapes is in column 1, SUN RGBD in column 2, and the Robot@Home Data is in column 3. The first row (a) shows the original images, the second row (b) shows our results from  the SceneNet RGB-D pretrained model, the third shows results from the ImageNet pretrained model, and the last row shows the ground truth.}
\label{fig:seg_image_collage}
\end{figure}

\section{Related Work} \label{sec:num2}

\subsection{Semantic Segmentation in Computer Vision}

Recent work on semantic segmentation using CNNs has greatly advanced the subfield, but with limited focus on robot vision. Most approaches often use some combination of large convolutions, a strictly serialized layer setup \cite{simonyan2014vgg}, networks with over one hundred layers, or fully connected layers at the end of the network \cite{DBLP:journals/corr/Garcia-GarciaOO17}. As a result, these methods have hundreds of millions of parameters, are slow to train, and evaluate far outside of real-time for semantic segmentation. Thus, models that can do segmentation in real-time, as is required for robotics, are sparse in the literature and while there has been some effort for real-time general vision architectures \cite{wu2017squeezedet}\cite{rastegari2016xnor}, they often perform poorly at segmentation. E-Net \cite{paszke2016enet} is the first effort to take advantage of all of these techniques, and the proposed architecture is specifically designed for real-time segmentation. None of these works, however, has examined the efficacy of the above architectures when trained on small, noisy robot vision datasets.

\subsection{Segmentation Datasets}

Segmentation datasets have become more relevant and abundant recently as the computer vision community has become more interested in segmentation, but they rarely apply to robotics directly and are not large enough to train models without pretraining. Popular segmentation datasets that are relevant to robotics include datasets for autonomous driving \cite{brostow2009semantic}\cite{geiger2012we}\cite{Cordts2016Cityscapes}, and indoor datasets \cite{Silberman:ECCV12}\cite{xiao2013sun3d}\cite{Janoch:EECS-2012-85}\cite{song2015sun}. Recently, the Robot@Home dataset \cite{ruiz2017robot} was published, which provides over 30K instance-labeled frames from over 80 sequences of a real robot navigating in six unique indoor environments. This is a vast improvement over previously available datasets for robot vision research, however all these datasets are still too small to be useful without pretraining and do not represent the complexities of robot vision well. 
 
Synthetic segmentation datasets have become popular recently, but are still underutilized and do not randomize the simulation conditions to increase diversity and remove bias. For autonomous driving, efforts including Shafaei \etal \cite{shafaei2016play}, Virtual KITTI \cite{gaidon2016virtual}, ``Driving in the Matrix" \cite{Johnson-Roberson:2017aa}, and the GTA dataset \cite{richter2016playing} all use high-fidelity simulations or video games to efficiently build realistic datasets, but all have less than 50K frames, which is many times too small to train robust autonomous vehicle perception systems. The SYNTHIA dataset \cite{ros2016synthia} contains 200K frames captured across eight cameras that form a 360-degree array, but this leaves only 25K examples to train systems with forward-facing cameras. For indoor environments, Song \etal \cite{song2016ssc} and Qui \etal \cite{qiu2016unrealcv} created datasets of over 2 million frames from thousands of rooms, but still are subject to dataset bias by being highly ordered. Taking this idea even farther, McCormac \etal created a comprehensive indoor dataset called SceneNet RGB-D \cite{mccormac2016scenenet} which generates and renders 5 million rendered RGB-D frames sampled from video through 3D scenes with randomized object compositions, textures, lighting, and camera trajectories. However, despite the availability of this large scale simulated data, computer vision researchers and roboticists alike continue to use ImageNet for pretraining since little evidence exists to properly explain the benefits of using simulated data. 

\subsection{Real-time Segmentation in Robot Vision}

There are some recent efforts catering to real-time semantic segmentation of objects for robotic systems, especially with regards to autonomous driving \cite{kostavelis2015semantic}\cite{Vertens2017SMSnetSM}\cite{Ha2017MFNetTR}\cite{zhang2016instance}\cite{teichmann2016multinet} and grasping \cite{Detry2017TaskorientedGW}\cite{DBLP:journals/corr/TobinFRSZA17}\cite{bousmalis2017using}; however the majority of these do not strictly enforce real-time requirements, nor do they utilize synthetic data. James \etal \cite{pmlr-v78-james17a} examine the effect of changing the amount of synthetic pretraining data for their grasping task, but the smallest amount of data they consider is 100K images, they do not operate in natural environments, nor are they concerned with real-time performance. Madaan \etal \cite{Madaan2017WireDU} and Lin \etal \cite{lin2017learning} utilize synthetic data to train a custom real-time CNN for segmentation on a robot. However, each of these works focus on simply segmenting a binary mask. Our work segments entire scenes into many objects, motivated by scenarios of more complex robots like home robots and autonomous cars. Most closely related to our work, McCormac \etal improves semantic segmentation for the task of depth-based simultaneous localization and mapping for robots \cite{mccormac2016semanticfusion} using synthetic image data. McCormac \etal \cite{mccormac2017beatimagenet} go on to demonstrate that the large U-Net \cite{ronneberger2015u} architecture pretrained on SceneNet RGB-D outperforms the same architecture pretrained on ImageNet. Our work differs from prior efforts in that it examines whether the use of synthetic data improves a small, real-time architecture, analyzes the ways in which using synthetic data affects performance as we vary the amount of target task data, and examine how high-level similarity between pretraining and target data effects performance. 

Closely related to this work is recent work on domain adaptation, including in semantic segmentation. In different task areas, the domain randomization work of \cite{DBLP:journals/corr/TobinFRSZA17} \etal illustrates how randomization helps with transfer and adaptation for robot learning from vision. Mayer \etal \cite{mayer2018makes} expands further on this notion in learning optical flow, observing that accuracy is not necessarily beneficial to domain adaptation from simulation. In segmentation, the aforementioned SYNTHIA \cite{ros2016synthia} dataset uses diverse simulation data as a means of on domain adaptation between different weather conditions in autonomous driving. Hoffman \etal \cite{Hoffman2016FCNsIT} and Zhang \etal \cite{zhang2017curriculum} take a slightly different tack, exploring methods of building domain adaptation capabilities into the learner itself, instead of depending on diverse training data. While these works are related, our work focuses specifically on the challenge of adaptation of small networks that lack representational capacity to distinguish signal from dataset bias, and how carefully selecting pre-training data can alleviate this issue and provide increasing degrees of improvement as fine-tuning data is reduced.

\section{Transfer Learning with Synthetic Data as Bias Reduction} \label{sec:num3}

Dataset bias is an often overlooked component of training data-driven computer vision models. Especially in the context of transfer learning, the assumptions made about which distributions are ``similar'' are often naive, and as a result leave performance gains unrealized. We first examine the distributions modeled to mathematically justify the use of a synthetic dataset over ImageNet data.

\subsection{Transfer learning and pretraining} \label{sec:num3_1}
The most effective and commonly utilized method for augmenting performance of models using small datasets is \textit{transfer learning} \cite{yosinski2014transferable}\cite{oquab2014learning}\cite{goodfellow2016deep}, where one exploits the similarity in two distributions, $P_{i}$ and $P_{t}$, by using parameters optimized to represent the initial distribution $P_{i}$ to provide an improved starting point for learning the target distribution $P_{t}$. 

Transfer learning is most often executed using \textit{pretraining}, where a model is first trained on one task with a similar input domain to the target task, and then that model is used to initialize the network parameters. There are two typical ways of using the pretrained model: either the target task data is used to continue training the entire model over the transfered parameters, or the pretrained parameters are ``frozen'', except for the inference parameters which are reinitialized randomly, and the target task data is used to optimize only the inference layers. 

In supervised learning, the CNN and its parameters $M_{\theta}$ can be thought of as forming an approximate representation of the task likelihood distribution $P_t(Y \vert X)$.  By Bayes Theorem, we know that 
\begin{equation}
M_{\theta} \approx P_{t}(Y \vert X) = \frac{P_{t}(Y) P_{t}(X \vert Y)}{P_{t}(X)}
\label{eq:nn_bayes}
\end{equation}
where the posterior distribution $P_{t}(X \vert Y)$ in this case is the generative task of constructing images given a label as input. 

In transfer learning, the assumption is that, for two supervised problems where the input is sampled from all natural images, the divergence between task likelihood distributions varies as the divergence between the input distributions:
\begin{equation}
D_f(P_{i}(X), P_{t}(X)) \sim D_f(P_{i}(Y \vert X),P_{t}(Y \vert X))
\label{eq:sim_approx}
\end{equation}
%
%

This assumption is justified by empirical successes improving a wide range of computer vision problems by first pretraining on ImageNet \cite{yosinski2014transferable}. These assumptions, however, do not account for dataset bias nor the difference between $P_{i}(Y)P_{i}(X \vert Y)$ and $P_{t}(Y)P_{t}(X \vert Y)$, or $\Delta P(Y)P(X \vert Y)$, and as such leave room for improvement.


\subsection{The transfer learning gap} \label{sec:num3_2}

The Torralba \etal \cite{torralba2011unbiased} study on dataset bias shows that even the most carefully constructed image datasets contain significant and distinct enough bias that a simple linear discriminative model can distinguish between them based on their inputs alone. \cite{torralba2011unbiased} concludes that dataset bias, specifically input bias, comes in three main forms: \textbf{selection bias} (images selected manually inherently have more bias than those obtained randomly or automatically), \textbf{capture bias} (image content is curated, e.g. photographs taken by people, objects are most often photographed from specific angles), and \textbf{negative set bias}, (datasets only collect items of interest, yielding models that do not sufficiently represent negative cases)\footnote{It is worth noting that negative set bias is less of a concern in segmentation because pixels are conditionally dependent on their neighborhood, and in general the more classes something is trying to predict the more negative examples that class has.}. Additionally, one bias that Torralba \etal do not mention is \textbf{annotation bias}, or the bias from errors in the human annotation; this is especially important in segmentation due to the large effect annotation errors near object boundaries have on prediction. Standard computer vision datasets like ImageNet have inputs that are carefully selected, and sampled from biased Web images that were captured and selected by the humans that took them, and have humans annotating them.

Robot vision datasets, on the other hand, are mostly uncurated, noisy, and have human annotators who bias segmentation annotations differently than the annotations of an image classification the task like ImageNet. We can assume, therefore, that in transferring features from a dataset like ImageNet there exist some difference in the biases of the inputs $P_(X)$ and annotations $P_(Y)$ that cannot be closed trivially. These three types of input bias, annotation bias, and $\Delta P(Y)P(X \vert Y)$ comprise the \textit{transfer learning gap} of two datasets, which we will call $G^{tr}$. 

Synthetic image datasets created with the same task as the target task can reduce $G^{tr}$ for the process of transfer learning. With camera angles and lighting generated randomly, these datasets have virtually no selection and capture bias, except for the capture bias in the construction and arrangement of the simulated scene. This can be further reduced by introducing some inherent randomness in its initialization. Simulation also has the benefit of generating perfect annotations for free with objectively no bias. Intuitively one might guess that transferring from the same task domain will improve the effectiveness of pretraining. We can justify this intuition by inspecting the decomposition of Equation \ref{eq:nn_bayes}; while the distribution $P_{t}(Y)P_{t}(X \vert Y)$ is difficult to observe, we can see that by rewriting Equation \ref{eq:nn_bayes} for a known image $ x \in X $, the target task distribution is proportional to the product of its label distribution and its generative posterior.
\begin{equation}
M \propto  P_{t}(Y) P_{t}(X=x \vert Y)
\label{eq:X=x}
\end{equation}

As a result, it can be stated that by attempting to match the target task as closely as possible in simulation, we can reduce $\Delta P(Y)P(X \vert Y)$. 


\subsection{Improving small models with simulated data} \label{sec:num3_3}

Given that synthetic datasets show the ability to reduce these four biases, plus $\Delta P(Y)P(X \vert Y)$, we can confidently predict that $G^{tr}_{ImNet} > G^{tr}_{Synth}$. However, simulations come with their own additional bias, referred to by the research community as the ``Sim2Real'' or the \textit{reality gap} \cite{DBLP:journals/corr/TobinFRSZA17}, $G^{s2r}$. Notably, no matter their fidelity and realism, simulations will always struggle to properly model sensor noise, imperfections, and complex physical phenomena. This bias manifests as a lack of what Zhou \etal call ``coverage'' in the data, defined as ``quasi-exhaustive representation of the classes and variety of exemplars'' \cite{zhou2017places}. Good coverage in synthetic datasets is generally achieved by having a wide range of random camera angles, lighting, textures, object arrangements, and additional noise \cite{DBLP:journals/corr/TobinFRSZA17}. We hypothesize that as long as the dataset compensates for the reality gap by giving sufficient coverage of the input distribution, we find that, 
\begin{equation}
G^{tr}_{ImNet} > G^{tr}_{Synth} + G^{s2r}
\label{eq:gap_ineq}
\end{equation}
As a result CNN pretraining with synthetic data will still be beneficial.

The shortcomings of transfer learning affect all CNNs, but not equally. During pretraining, if bias exists in the pretraining dataset, large models with many parameters have the capacity to model both the underlying distribution and the bias. Small models, on the other hand, do not have the capacity to model both, and over time will tend toward to learning the bias because it gives the lowest loss. As a result, large models will transfer the underlying distribution during fine-tuning and be able to ignore pretraining bias, while small models will have to unlearn the effects of the pretraining bias. Therefore, we hypothesize that for transfer learning, pretraining on data that has less bias with respect to the target data will show greater performance improvement with small models. Comparing our results to that of McCormac \etal \cite{mccormac2016scenenet} validates this hypothesis.

\begin{figure}
\centering
    \subfigure[This plot shows the mean IoU performance of models pretrained on SceneNet RGB-D and ImageNet as a function of the total amount of fine-tuning data a pretrained model is trained with, scaled logarithmically. Results for models pretrained on SceneNet RGB-D are in blue and results for models pretrained on ImageNet are in red.]{\includegraphics[width = \columnwidth]{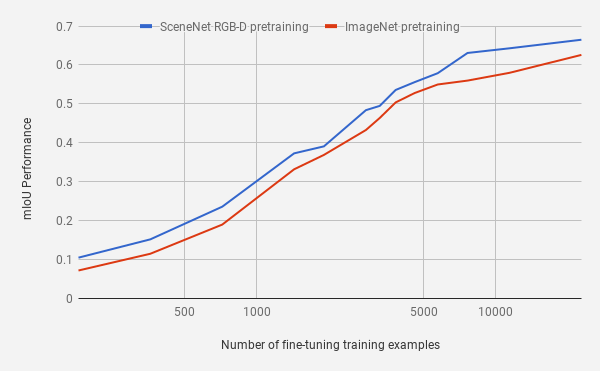}}
    \subfigure[This plot shows the percent improvement of the SceneNet RGB-D pretrained models over the ImageNet pretrained models for a given percentage of the total fine-tuning data. We can see that pretraining using synthetic segmentation data generally gives more improvement for smaller quantities of fine-tuning data]{\includegraphics[width = \columnwidth]{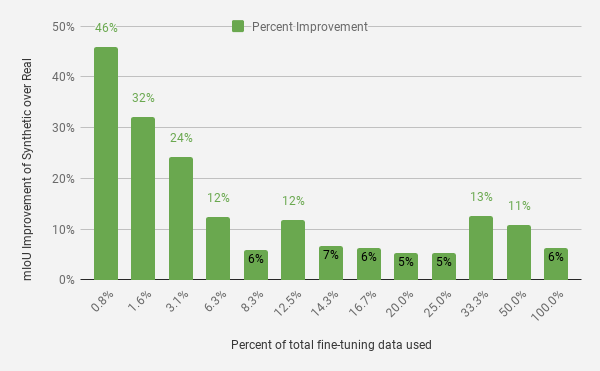}} 
\label{fig:robo_charts}
\caption{Chart of results from the Robot@Home training size variation experiment. It is interesting to note that the SceneNet RGB-D pretrained models always outperforms the ImageNet pretrained model for all the different training set sizes. }
\end{figure}

\section{Approach} \label{sec:num4}

We conducted three experiments to validate that Equation \ref{eq:gap_ineq} holds for small CNNs trained with sparsely supervised robot perception data: (1) An ablation experiment to demonstrate that using synthetic data for pretraining (compared to ImageNet) improves real-time models on standard semantic segmentation datasets, (2) a data withholding experiment to compare the models pretrained with synthetic data to the models pretrained with ImageNet by measuring their performance on a held out set of robot perception data after being fine-tuned using increasingly restricted amounts of robot perception data, and (3) a high-level similarity experiment to demonstrate that the high-level task similarity between pretraining and fine-tuning datasets has only a minor effect on model performance compared to the effects of bias reduction discussed in Section \ref{sec:num3}.

In this section, we outline how we selected a synthetic dataset that met the requirements outlined in Section \ref{sec:num3_2}, standard datasets relevant to robotics, preferably one for autonomous driving and one for an indoor scenario as those are two domains in which robots can benefit from segmentation, a robot dataset for the second experiment, and a semantic segmentation CNN architecture that could run in real-time on a robot.

\subsection{Dataset Selection} \label{sec:num4_1}

For our standard segmentation datasets, we use the SUN RGB-D \cite{song2015sun} indoor dataset and the Cityscapes \cite{Cordts2016Cityscapes} autonomous driving dataset. We selected SUN RGB-D because it has 37 challenging semantic classes and it is one of the largest real semantic segmentation datasets for indoor environments. We selected Cityscapes as it is the most recent real autonomous driving dataset, has 19 semantic classes which is more than most driving datasets, and 5K frames with ``fine" annotations from a series of driving videos.

For our primary synthetic pretraining dataset, we used SceneNet RGB-D \cite{mccormac2016scenenet}. SceneNet RGB-D has over 5 million photorealistic training images sampled from randomized smooth trajectories through 16K room configurations. Each unique room configuration has a random set of contextually-relevant objects initialized with both random pose and texture, and random lighting, and all frames are intentionally perturbed with realistic noise. Moreover, the dataset is labeled instance-wise, so we opted to mapping the objects to 13 classes, which made it more adaptable as a pretraining model for transfer learning and consistent with the experiments of McCormac \etal \cite{mccormac2017beatimagenet}. The diversity of the data in this dataset makes it especially well suited for large-dataset pretraining. 

Lastly, for our robot data we use the Robot@Home dataset \cite{ruiz2017robot}. Created with the intention of semantic mapping by a household robot, Robot@Home was collected over four years by recording 81 video sequences of a robot driving around 36 unique unstructured human spaces. The robot was equipped with four Primesense RGB-D cameras for recording the visual frames and a 2D laser scanner to improve mapping capabilities. Objects instance segmentation labels are provided for 32937 frames across 72 sequences, which like SceneNet RGB-D, has mappings to standard object segmentation class labels. We chose to use the mapping to the original 41 SUN categories to increase the difficulty of this last experiment.

\subsection{Model Architecture} \label{sec:num4_2} 

The architecture best suited to our task is the E-Net \cite{paszke2016enet} architecture. E-Net is an encoder-decoder style CNN composed mainly of ``bottleneck" modules, which combine three convolutions and a skip layer. The E-Net design combines an initial network-in-network module with quick downsampling to strike a compromise between reducing the number of parameters in the network and the representational power of low-level features from layers close to the original image resolution. The remainder of the network is 16 bottleneck modules for the encoder, and 5 such modules for the decoder. Unlike most segmentation networks, E-Net has an asymmetric encoder-decoder design as the encoder is the main feature extractor and the decoders are often responsible for using large quantities of parameters. Our implementation of E-Net runs very efficiently: over 56 frames per second on average for a 256x256 input on a single NVIDIA GTX 1080Ti GPU.

\begin{table}
\renewcommand{\arraystretch}{1.3}
\caption{Ablation Study Metrics - Pixel Accuracy}
\label{table:abl_acc}
\centering
\begin{tabular}{|c||c|c|}
\hline
& \multicolumn{2}{c|}{\bfseries Real Dataset} \\
\hline
\bfseries Pretraining & \bfseries Cityscapes & \bfseries SUN RGB-D \\
\hline\hline
No pretraining & 0.851 & 0.483 \\
\hline
ImageNet & 0.860 & 0.516 \\
\hline
SceneNet RGB-D & \bfseries 0.863 & \bfseries 0.585 \\
\hline
\end{tabular}

\renewcommand{\arraystretch}{1.3}
\caption{Ablation Study Metrics - Mean Accuracy}
\label{table:abl_mAP}
\centering
\begin{tabular}{|c||c|c|}
\hline
& \multicolumn{2}{c|}{\bfseries Real Dataset} \\
\hline
\bfseries Pretraining & \bfseries Cityscapes & \bfseries SUN RGB-D \\
\hline\hline
No pretraining & 0.410 &  0.200  \\
\hline
ImageNet &  0.501 & \bfseries 0.368 \\
\hline
SceneNet RGB-D & \bfseries 0.585 & 0.346 \\
\hline
\end{tabular}

\renewcommand{\arraystretch}{1.3}
\caption{Ablation Study Metrics - mean IOU}
\label{table:abl_iou}
\centering
\begin{tabular}{|c||c|c|}
\hline
& \multicolumn{2}{c|}{\bfseries Real Dataset} \\
\hline
\bfseries Pretraining & \bfseries Cityscapes & \bfseries SUN RGB-D \\
\hline\hline
No pretraining & 0.346 &  0.118\\
\hline
ImageNet & 0.392 &  0.180 \\
\hline
SceneNet RGB-D & \bfseries 0.489 & \bfseries 0.227 \\
\hline
\end{tabular}
\end{table}

\section{Experiments and Results} \label{sec:num5}
We compare models trained from scratch in two different ways. In the first case, a model is pretrained on an image classification task using the ImageNet dataset. As ImageNet is an image classification dataset it is only used to pretrain the encoder --- standard practice for transfer learning from classification to segmentation. The model is then fine tuned on the target dataset with randomly initialized output layers, which in the case of segmentation is the ``decoder''. In the second case, a model is pretrained end-to-end on a semantic segmentation task using the SceneNet RGB-D dataset, and then the entire model is fine tuned with the target dataset.
\begin{table*}[t]
\renewcommand{\arraystretch}{1}
\caption{Robot@Home mIoU Dataset Variance Evaluation}
\centering
\title{This table shows the performance difference in the two pretrained models given different amounts of fine-tuning data.}
\label{table:RobotatHome_mIoU}
\begin{center} 
\scalebox{0.79}{
\begin{tabular}{ |c||c|c|c|c|c|c|c|c|c|c|c|c|c| }
\hline
 &  \multicolumn{13}{c|}{ \bfseries Percentage of Full Robot Fine-tuning Training Set (number of examples)}  \\
\hline
\bfseries Pretraining & 0.8\% & 1.6\% & 3.1\% & 6.3\% & 8.3\% & 12.5\% & 14.3\% & 16.7\% & 20.0\% & 25.0\% & 33.3\% & 50.0\% & 100.0\% \\
\bfseries Dataset & (179) & (358) & (717) & (1,434) & (1,912) & (2,868) & (3,278) & (3,824) & (4,589) & (5,736) & (7,648) & (11,473) & (22,946) \\
\hline\hline
ImageNet &  0.072 & 0.115 & 0.19 & 0.332 & 0.369 & 0.433 & 0.464 & 0.504 & 0.528 & 0.55 & 0.56 & 0.58 & 0.626 \\
\hline
SceneNet RGBD&  0.105 & 0.152 & 0.236 & 0.373 & 0.391 &  0.484 & 0.495 & 0.536 & 0.556 & 0.579 & 0.631 & 0.643 & 0.665 \\
\hline
\% Improvement &  45.8\% & 32.2\% & 24.2\% & 12.3\% & 6.0\% & 11.8\% & 6.7\% & 6.3\% & 5.3\% & 5.3\% & 12.7\% & 10.9\% & 6.2\% \\
\hline
\end{tabular}}
\end{center}
\end{table*}

\subsection{Implementation Details} \label{sec:num5_1}
We implemented E-Net using the \textit{PyTorch} \cite{paszke2017automatic} framework \footnote{All code and models can be found on our github \href{https://github.com/balloch/synth-seg}{here}}. The network is trained using negative log linear loss on a Softmax function and optimized using the adaptive gradient descent algorithm, Adam \cite{kingma2014adam}. A brief hyper-parameter search was conducted on the initial learning rate $\alpha \in [10^{-2}, 5*10^{-3}, 10^{-3}, 5*10^{-4}, 10^{-4}]$ and we found that the initial learning rate of $\alpha=1e^{-3}$ was good for both training from scratch and fine tuning. For the other hyper-parameters, we used the values suggested by Kingma \etal. We also experimented with mini-batch sizes $b \in [10,32,64,128]$ for training, and found that the results were fairly similar, with 128 converging the most efficiently for the pretraining datasets. The real, non-ImageNet datasets were trained with a batch size of 32.  For weight initialization we randomly sample from a Gaussian $= N(\mu=0,\sigma=0.02)$ distribution. All images were scaled to a resolution of 256x256 for our experiments. Training was performed on NVIDIA K40 Quadro, NVIDIA TITAN X, and NVIDIA GTX 1080Ti GPUs.

In each experiment, the final dataset was evaluated on three metrics standard to semantic segmentation. These are pixel accuracy, mean accuracy, and the mean Intersection over Union (mIoU) measure. For $M$ total classes, and for some predicted class $j$, $m_{ij}$ are the number of pixels in class $i$ that are predicted to be in class $j$. The pixel accuracy measures the ratio of pixels predicted correctly to all labeled pixels; this is a good indicator of how well the segmentation did relative to random chance. The mean accuracy measures the average across classes of the ratios of pixels predicted correctly to the total number of pixels in a label class. This measures the accuracy of the assignment over all classes. Lastly, mean IoU measures the average across classes of the the ratios of pixels predicted correctly to the total number of pixels in a label class plus the number of pixels in the prediction class that were not correctly classified. This is the most stringent measurement, and the best indicator of model performance in practice. 





\subsection{Ablation Experiment} \label{sec:num5_2}

In this experiment, we used each of the standard datasets, SUN RGB-D and Cityscapes, to train a model from scratch, to fine tune over ImageNet pretraining, and to fine tune over SceneNet RGB-D pretraining. In addition to the two regular training paradigms, in this experiment, models were trained for each target dataset using just the target training data (from-scratch) and for consistency with standard research practices this dataset was augmented using resizing and horizontal flipping for all three training scenarios. For Cityscapes, the predesignated train-val-test splits on the ``fine" annotations were used. For SUN RGB-D, which does not have predetermined splits, the 10335 images were randomly sampled and split into a $70\%-10\%-20\%$ train-val-test split. It was observed that at a batch size of 32 it would take roughly 50 epochs for convergence of training with no loss of validation performance (i.e. early stopping).

For the fine-tuning process over ImageNet, the decoder was initialized in the same manner as the network when training from scratch, and all datasets were trained upon with validation monitoring for early stopping (around 30 epochs). For the fine-tuning process over SceneNet RGB-D, only the decoder was trained with validation showing the models converging at roughly 30 epochs. Mirroring the process of the from-scratch training, SUN RGB-D and Cityscapes were used to fine tune both pretrained models at a batch size of 32, validating after each epoch with early stopping.

Tables \ref{table:abl_acc}-\ref{table:abl_iou} show the results of the ablation experiment. In the ablation experiment, the results show that for fine-tuning and testing on Cityscapes, the SceneNet RGB-D pretrained model outperforms the ImageNet pretrained model by $0.43\%$ in pixel accuracy, $16.68\%$ in mean accuracy, and $24.85\%$ in mean IoU. For fine-tuning and testing on SUN RGB-D  the SceneNet RGB-D pretrained model outperforms the ImageNet pretrained model by $13.37\%$ in pixel accuracy and $26.34\%$ in mean IoU. For SUN RGB-D, the ImageNet pretrained model outperformed the SceneNet RGB-D pretrained model by $5.96\%$ in mean accuracy; however models that perform well in mean accuracy and not as well in mean IoU learn to over-fit to a subset of the most heavily represented classes in the dataset, which is consistent with our hypothesis that a model will have a harder time training over the biases of ImageNet. These results validate the hypothesis that using synthetic data to pretrain real datasets is still a viable approach for architectures like E-Net that have far fewer parameters than typical segmentation networks. Furthermore, this confirms the findings of McCormac \etal and demonstrates that their conclusions extend beyond large parameter networks. Interestingly, the ablation results shown here are more dramatic than in McCormac \etal in spite of the fact that they pretrained for longer on a larger model. This validates our hypothesis from Section \ref{sec:num3_3} that transfer learning from synthetic data is more effective for smaller networks.

\subsection{Robot@Home Dataset Experiment} \label{sec:num5_3}

Like SUN RGB-D, the Robot@Home dataset did not have predetermined splits, so following our methods with the ablation study, the 32937 images were randomly sampled into a $70\%$-$10\%$-$20\%$ train-val-test split. 

The purpose of this experiment is to examine how the efficacy of transfer learning changes with fine tuning robot datasets of a variety of sizes. To examine the effects, we down-sample the training split into other small training sets. Specifically, keeping the validation and test sets untouched and unchanged, we create additional training datasets that are $\frac{1}{2}$, $\frac{1}{3}$, $\frac{1}{4}$, $\frac{1}{5}$, $\frac{1}{6}$, $\frac{1}{7}$, $\frac{1}{8}$, $\frac{1}{16}$, $\frac{1}{32}$, $\frac{1}{64}$, and $\frac{1}{128}$ of the size of the original 22,946 training set (see Table\ref{table:RobotatHome_mIoU} for more details). With those sub-sampled fine-tuning datasets, an ImageNet pretrained model and a SceneNet RGB-D pretrained model is fine-tuned for each of the 12 training datasets. 

For the Robot@Home dataset, the evaluations for the two types of models show very interesting results. In Figure \ref{fig:robo_charts} we show the mean IoU of the SceneNet RGB-D and ImageNet models as a function of what fraction of the dataset they were trained on. SceneNet's best model, trained on the full set of training data, outperformed ImageNet's best model by 15.6\% in mean IoU. The model pretrained with SceneNet outperforms ImageNet for every data subdivision; even more interestingly, the performance difference is such that in most cases the SceneNet pretrained model requires between anywhere from $15\%$ to $50\%$ less finetuning data than the ImageNet model to match its performance. This is an especially meaningful result because it shows that roboticists considering the time and monetary investment of acquiring and labeling more data may want to first consider investing time in sampling data from a simulation before expensively collecting more real world data.

\subsection{High-Level Similarity Experiment} \label{sec:num5_3}

To explore the effects of high-level similarity between pretraining and target task datasets, the third experiment compares results of models pretrained on different synthetic data on real segmentation datasets. We ran evaluations for a four-way cross comparison to test if high-level domain similarity in two datasets impacts training, looking at the indoor navigation and autonomous driving datasets for both synthetic and real data. 

For this experiment, the GTA dataset\cite{richter2016playing} is used as the autonomous driving pretraining data. This dataset has 25K densely annotated frames sampled from the video game Grand Theft Auto (GTA), and while 25K is small for a pretraining dataset, is was sufficient for the purpose of the experiment. To make a more apt comparison, we sub-sampled a 25K training set from SceneNet RGB-D, which we refer to as SceneNet RGB-D (25K). This was used as the indoor navigation pretraining dataset. The four-way cross comparison therefore was:
\begin{itemize}
 \item SUN RGB-D pretrained on SceneNet RGB-D (25K) (similar)
 \item Cityscapes pretrained on GTA (similar)
 \item SUN RGB-D pretrained on GTA (not similar)
 \item Cityscapes pretrained on SceneNet RGB-D (25K) (not similar)
\end{itemize}

\begin{table}[t]
\renewcommand{\arraystretch}{1.3}
\caption{High-Level Similarity - mean IoU Comparison}
\label{table:similarity}
\centering
\begin{tabular}{|c||c|c|}
\hline
& \multicolumn{2}{c|}{\bfseries Real Dataset} \\
\hline
\bfseries Pretraining & \bfseries Cityscapes & \bfseries SUN RGB-D \\
\hline\hline
GTA  & 0.467 &  0.205 \\
\hline
SceneNet RGB-D (25K) & 0.379 & 0.193 \\
\hline
SceneNet RGB-D & \bfseries 0.489 & \bfseries 0.227 \\
\hline
\end{tabular}
\end{table}

When comparing the models pretrained on the 25K synthetic image datasets, the ``high-level domain similarity'' pairs i.e. Cityscapes trained over the GTA dataset and SUN RGB-D trained over SceneNet RGB-D (25K), achieved mIoU scores of $0.467$ and $0.193$ respectively. It is worth noting that even though these datasets are considered to be typically far too small to be used for pretraining, these mean IoU scores are greater than those achieved by Cityscapes and SUN RGB-D models pretrained on ImageNet. For the other two cases, Cityscapes trained over SceneNet RGB-D (25k) achieved a score of $0.379$ mIoU and SUN RGB-D trained over GTA achieved a score of 0.205 mIoU. The results of this experiment reinforce our hypothesis that pretraining data with high-level similarity has some positive effect on performance.

It is worth noting that the SUN RGB-D model trained over GTA performed better even though the dataset domains are semantically less similar, which indicates that high-level domain similarity may not help in all cases. These results show that for two synthetic pretraining datasets of the same size from different semantic domains, models may perform better if they are pretrained on data that is similar to their goal domain. However, neither 25K frame dataset gave better performance in this experiment than the Cityscapes and SUN RGB-D models trained over the full SceneNet RGB-D training set, as can be seen in Table \ref{table:similarity}, which is further consistent with our hypothesis that for two synthetic datasets that address the same task and sample their input from the same domain, a separate factor, in this case dataset size, dominates the other, smaller differences that affect $G^{tr}_{Synth}$.

\section{Conclusion} \label{sec:num7} 

In this work, we investigated the potential gains of using synthetic data to augment the training process of small CNNs designed for real-time semantic segmentation in robots with small target training sets. We compared the improvements afforded by synthetic data to traditional data augmentation and transfer learning from image classification. The performance gains from pretraining with synthetic data indicate that the degree to which this closes the transfer gap $G^{tr}$ for these models is reliably greater than the bias introduced by the ``Sim2Real'' problem, $G^{s2r}$. We also documented the evolution of improvements to real-time semantic segmentation models as access to real data decreases, and showed that as dataset size decreases, the improvements from using task similar synthetic data increase exponentially compared to ImageNet. 

We are currently considering how this technique might be used with other solutions to semi-supervised or weakly supervised problems, and whether there are other, more interesting ways to effectively use simulation to improve real-time segmentation, potentially as a feedback signal for model architecture search, or even more interestingly as an oracle in an active vision or lifelong learning agent.



{\small
\bibliographystyle{ieee}
\bibliography{references,IEEEabrv}
}

\end{document}